\documentclass[conference]{IEEEtran}
\IEEEoverridecommandlockouts
\usepackage{cite}
\usepackage{amsmath,amssymb,amsfonts}
\usepackage{algorithmic}
\usepackage{graphicx}
\usepackage{textcomp}
\usepackage{xcolor}
\usepackage{balance}

\def\BibTeX{{\rm B\kern-.05em{\sc i\kern-.025em b}\kern-.08em
    T\kern-.1667em\lower.7ex\hbox{E}\kern-.125emX}}

\begin{document}

\title{Air Signing and Privacy-Preserving Signature Verification for Digital Documents}

\author{
    \IEEEauthorblockN{
        P.~Sarveswarasarma, 
        T.~Sathulakjan, 
        V.~J.~V.~Godfrey, 
        Thanuja~D.~Ambegoda
    }
    \IEEEauthorblockA{
        Department of Computer Science and Engineering, University of Moratuwa, Sri Lanka\\
        Emails: \{sarveswarasarma.18, sathulakjan.18, victorjeyakumar.18, thanuja\}@cse.mrt.ac.lk
    }
}

\maketitle

\begin{abstract}
This paper presents a novel approach to the digital signing of electronic documents through the use of a camera-based interaction system, single-finger tracking for sign recognition, and multi commands executing hand gestures. The proposed solution, referred to as "Air Signature," involves writing the signature in front of the camera, rather than relying on traditional methods such as mouse drawing or physically signing on paper and showing it to a web camera. The goal is to develop a state-of-the-art method for detecting and tracking gestures and objects in real-time. The proposed methods include applying existing gesture recognition and object tracking systems, improving accuracy through smoothing and line drawing, and maintaining continuity during fast finger movements. An evaluation of the fingertip detection, sketching, and overall signing process is performed to assess the effectiveness of the proposed solution. The secondary objective of this research is to develop a model that can effectively recognize the unique signature of a user. This type of signature can be verified by neural cores that analyze the movement, speed, and stroke pixels of the signing in real time. The neural cores use machine learning algorithms to match air signatures to the individual's stored signatures, providing a secure and efficient method of verification. Our proposed system does not require sensors or any hardware other than the camera.
\end{abstract}

\begin{IEEEkeywords}
Air signing, Finger Tracking, Hand gestures, Computer Vision, One-Shot, Siamese Network
\end{IEEEkeywords}

\section{Introduction}
Many application sections use desktop interfaces that work with gestures. But gestures are still a recent trend. Computer vision is a tool that can aid with this. It makes use of the camera to examine the pictures and make objects, people, or events identifiable. Current attempts to solve the problem of gesture recognition include the use of mechanical devices and gloves, which require the user to wear a device and typically make use of loads of cables that connect the device to a computer. However, this is not practical for a low-quality camera like a webcam. Instead of using these gadgets, engagement strategies based on cameras can be used. Here, with 94\% True Positive (TP) and 85\% True Negative, we present a practical method to track a single finger using the camera as a sensor, recognize signs written in the air, and utilize the usual web camera in unique ways (TN). The next stage is to make it useful for instructing our model with hand gestures and postures to enhance HCI. They had only created an approach in [8] to observe fingertips on a standard webcam. Finding drawing lines to detect, track, identify, and convert into signable format (.png) or copy to the clipboard is a difficult challenge. While the handwritten text on paper and optical character recognition (OCR) technologies have both been widely investigated for identifying text in printed documents, applying these methods directly to detect a person's signature in the air is difficult and cannot, therefore, be generalized. Inconsistent lighting, block lettering, background noise, and visual distortions are further problems that can make it difficult to recognize signatures or motions in the air.

In an increasingly digital world, the need for secure and trustworthy document management has become paramount. Digital signing on PDFs has emerged as a widely adopted method for signing and exchanging electronic documents. However, ensuring the authenticity, integrity, and non-repudiation of these digitally signed documents remains a critical challenge. Signature verification plays a pivotal role in addressing these challenges by confirming the validity of digital signatures and providing assurance that the signed PDFs have not been tampered with or forged. This research paper aims to explore the significance of signature verification in digital signing on PDFs, examining its role in establishing trust, maintaining document integrity, enabling non-repudiation, and meeting regulatory and compliance requirements. Through a comprehensive analysis of the underlying principles, techniques, and benefits. The proposed remedy offers a revolutionary method of signing documents online called "Air Signature," and the necessity of robust signature verification mechanisms in the digital signing ecosystem, offering insights for practitioners, policymakers, and researchers in the field of secure document management.

\section{Related Work}
Identification of specifically marked fingertips in a scene led to the development of an early and simple fingertip detection technology. One effective method of detecting fingertips is to paint them directly and find the paint in a video can. Because paints are located without considering the color or geometric details of the hand, this is regarded as an indirect way of fingertip detection. Additionally, it is inconvenient since each time detection is performed, a user must paint their fingertips \cite{b1}. The HCI system Nakamura et al. set up \cite{b10} involved attaching LED light devices to fingertips, searching for the locations of LED light sources, and then treating the source positions as fingertips. They quickly acquired spatial knowledge about fingertips in this way.

Recently, based on a wearable camera, Sajid et al. \cite{b12} proposed a robust fingertip-tracking system that addresses the inconveniences caused by data gloves and can be used for signature authentication. However, the head-mounted wearable devices used in that paper are extremely expensive.

In \cite{b6,b7,b9} a limited amount of work has been done at a high level without having an external device. G. Tofighi et al. \cite{b6} show the step-by-step approach to detecting the fingertip using only a web camera. They proposed a novel approach to detecting pointing vectors in the 2D space of a room in live video from a common webcam. After background subtraction, the face and forehead are detected. The research was an experimental algorithm to extract the fingertips using the Freeman chain code algorithm. Background subtraction, face detection skin histogram template, hand detection, and contour processing for pointing gesture detection. By using the COG method identify the fingertips. This reference is useful because it helps to background subtraction from complex environments. V. Joseph et al. \cite{b9} proposed a convex shape constructed by a convex hull algorithm. By using this they identify fingertips then find coordinates of the extreme top as mentioned above and plot the coordinates continually. However, these methods do not have a combination of hand postures and gestures to control the sketches.

Therefore, a novel approach for Air Signature in a complex environment with no special devices is necessary.

In \cite{b17}, for the signature verification process, they used Convolutional Neural Networks (CNN) to learn features from the pre-processed genuine signatures and forged signatures. The CNN used is inspired by Inception V1 architecture(GoogleNet). The architecture uses the concept of having different filters on the same level so that the network would be wider instead of deeper. For training the model they have used three different publicly available datasets:

\begin{itemize}
    \item CEDAR: It contains signatures of 55 signers having a place with a different social and expert background.
    \item BHSig260: This dataset contains Hindi and Bengali language signatures. This has 100 signers from Bengali and around 160 from Hindi therefore a total of 260 signers.
    \item UTSig: UTSig dataset consists of signatures signed in the Persian language and contains a total of 8280 signatures. There are 115 individuals whose signatures are recorded.
\end{itemize}

Before training the model, they performed some pre-processing of these signature images. In the processing section, two images from the original signature are generated which are fed to the network.

1) Filtered Signature: The image is transformed into a grayscale image for better understanding since color doesn’t matter in the signature. The image is passed through a dilate filter and then through a Gaussian blur to reduce the noise.

2) Grayscale Signature: Similar to the filtered signature, the image is transformed into a grayscale image for better understanding.

They evaluated the model based on Accuracy, False Acceptance Rate (FAR), and False Rejection Rate (FRR). Here are some results of their work:

\begin{figure}[ht]
    \centering
    \includegraphics[width=0.4\textwidth,keepaspectratio]{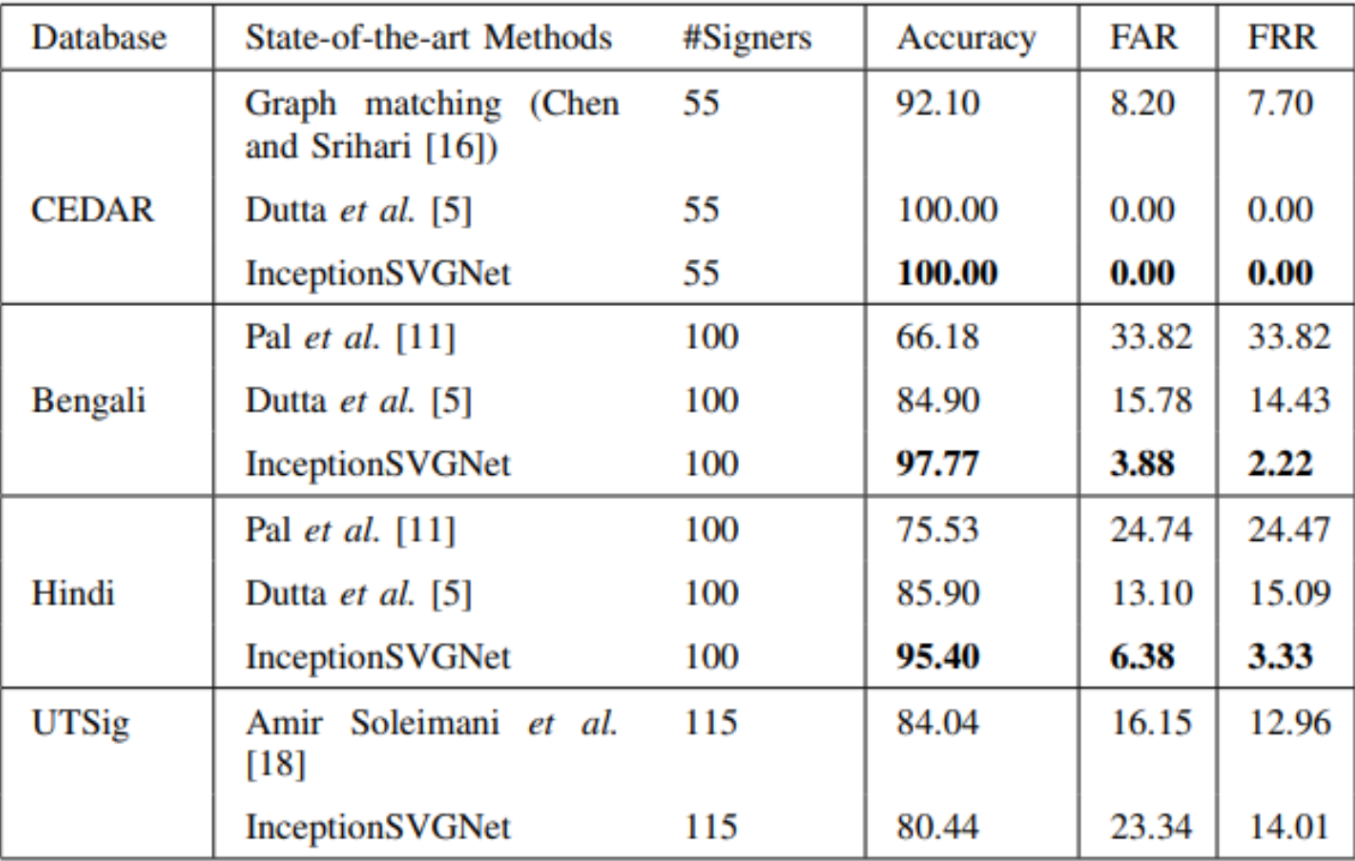}
    \caption{Datasets}
    \label{fig:datasets}
\end{figure}

The paper \cite{b18} "Siamese Neural Networks for One-shot Image Recognition" is a significant contribution to the field of one-shot learning. The paper introduces a new method that achieves state-of-the-art results on one-shot image classification tasks. The paper also discusses the potential applications of Siamese neural networks for one-shot image recognition.

The paper has been well-received by the research community. It has been cited over 1,000 times and has been used as a basis for several follow-up studies.

Here are some of the related work that has been inspired by the paper:

Prototypical Networks (Vinyals et al., 2016) are a type of Siamese neural network that uses a prototype representation for each class. Prototypical Networks have been shown to achieve state-of-the-art results on a variety of one-shot learning tasks.

MAML (Sung et al., 2017) is a meta-learning algorithm that can learn to adapt to new tasks with few examples. MAML is effective for one-shot image classification and other one-shot learning tasks.

Relational Siamese Networks (Chen et al., 2018) use a Siamese neural network to learn the relationships between pairs of images. Relational Siamese Networks are effective for one-shot image classification and other one-shot learning tasks.

The paper \cite{b16} has had a significant impact on the field of one-shot learning and has paved the way for several new and innovative methods.

\begin{figure}[ht]
    \centering
    \includegraphics[width=0.4\textwidth,keepaspectratio]{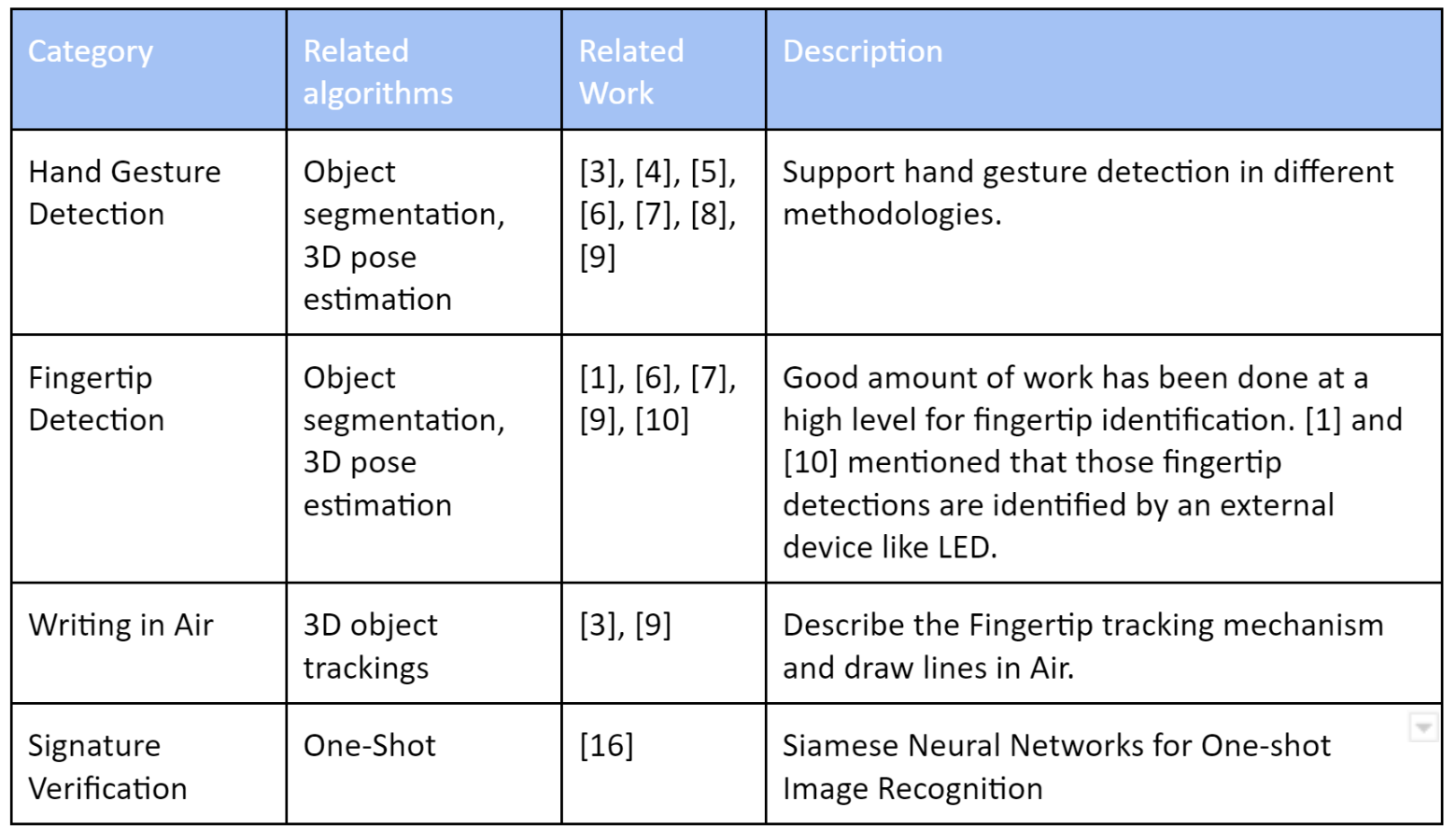}
    \caption{Table 2}
    \label{fig:table2}
\end{figure}

\section{Theoretical Background}
\subsection{Digital Signatures}
Digital writing signatures involve the use of a camera-based interaction system to capture and interpret the movements of a person's hand while writing in the air. The theoretical background of digital writing signatures encompasses several key components, including camera-based tracking, handwriting recognition, and motion analysis. The camera-based tracking system captures the movement of the person's hand in real-time, using computer vision techniques to track the trajectory and position of the hand. Handwriting recognition algorithms are then employed to analyze the captured movement and recognize the written characters or symbols. Motion analysis techniques further enhance the accuracy and robustness of the system by extracting relevant features from the captured data and identifying specific patterns or gestures. The theoretical foundations of digital writing signatures enable the development of intuitive and interactive systems for applications such as virtual reality interfaces and digital communication \cite{b19,b20}.

\subsection{Privacy Preserving Mechanisms}
Privacy-preserving mechanisms are techniques that allow users to share data without revealing their personal information. Privacy-preserving mechanisms are often used in conjunction with digital signatures to ensure that only authorized users can verify signatures.

\subsection{Air Signing Process}
Air signing is a theoretical concept that involves using gesture recognition technology to interpret hand movements and gestures performed in mid-air. The theoretical background of air signing encompasses computer vision, machine learning, and pattern recognition techniques. Computer vision techniques are used to capture and analyze video input from cameras, allowing the system to detect and track hand movements in real-time. Machine learning algorithms are then trained on annotated datasets to recognize specific hand gestures and classify them accordingly. Pattern recognition algorithms process the captured hand gestures and map them to predefined commands or actions. The theoretical foundations of air signing enable the development of accurate and real-time hand gesture recognition systems that can be applied in various domains, such as human-computer interaction, sign language recognition, and virtual reality applications.

\begin{itemize}
  \item The hand and the fingertip are difficult to distinguish from the complicated surroundings.
  \item Signature styles differ from one user to another. Some users may have a chance to sign without any continuity and some are not.
  \begin{itemize}
    \item The most frequent issue in a digital signature setting is hand tremors. So, the product has a responsibility to smooth that kind of signature.
    \item The camera fps put a huge impact on the output of the signature. When the user signs quickly, the result will either become a straight line or a discontinuous line.
  \end{itemize}
\end{itemize}

\section{Proposed Method}
\subsection{Air Signing}
Writing the signature in the air is the name of the initiative we're putting out as an alternative to digital signing. We have suggested a method for the digital sign on the air in front of the web camera utilizing the index finger after carefully examining our suggested solutions. The system needs to identify the fingertip of the hand as a first step, hence many methods are employed to do this. To accomplish this, our suggested method includes three steps: background subtraction, hand detection, and fingertip detection utilizing the Mediapipe library.

Because it is far simpler to estimate the bounding boxes of stiff objects like palms and fists than it is to identify hands with movement fingers, Mediapipe Library allows for the training of a palm detector rather than a hand detector to detect hands. In two-hand self-occlusion situations like handshakes, the non-maximum suppression strategy also works effectively because palms are smaller objects. Additionally, square bounding boxes that ignore other aspect ratios can be used to approximate palms, which leads to a reduction of 3-5 anchors. An encoder-decoder feature extractor is employed to offer context awareness for larger scenes, even for little objects. Finally, we decrease focus loss during training to accommodate a large number of anchors originating from the high-scale variance. Our second-hand landmark model uses regression, or direct coordinate prediction, to precisely localize the key points of 21 3D hand-knuckle coordinates inside the observed hand regions after recognizing the palm over the entire image. Self-occlusions and partially visible hands do not affect the model's ability to establish a trustworthy internal hand position representation. To create ground truth data, 30K genuine images were painstakingly annotated with 21 3D locations, as shown below. Additionally, to better cover the range of potential hand positions and provide additional oversight on the nature of hand geometry, render a high-quality synthetic hand model over a variety of backgrounds and map it to the related 3D coordinates.

\begin{figure}[ht]
\centering\includegraphics[width=0.75\linewidth]{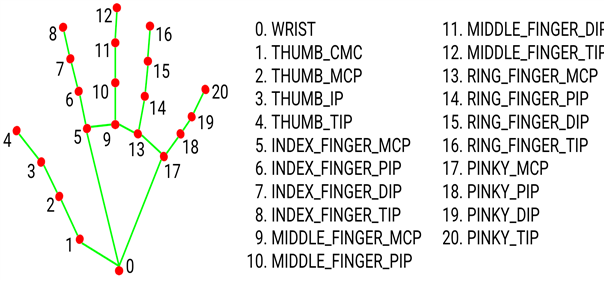}
\caption{Hand Landmarks.}
\label{fig:handlandmarks}
\end{figure}

As indicated in the above image \cite{b4,b8,b12,b16,b20}, the Y coordinate value of each fingertip's index is used to determine the up fingers, and the fingertip with the lowest value index is used to draw the pointer. Then, we utilize two hand postures to stop sketching the signature in the air and clear the signature on the canvas.

\begin{figure}[ht]
    \centering
    \includegraphics[width=0.1\linewidth,height=2cm]{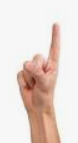}
    \hspace{0.5cm}
    \includegraphics[width=0.1\linewidth,height=2cm]{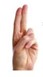}
    \hspace{0.5cm}
    \includegraphics[width=0.1\linewidth,height=2cm]{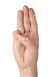}
    \caption{Drawing postures (Active, Stop, Erase)}
    \label{fig:postures}
\end{figure}

The system then records the final drawing that was drawn and turns it into a signature image using one of the usable image extensions (png). The user can then use that signature image anywhere the user wants to sign documents electronically.

\begin{figure}[ht]
    \centering
    \includegraphics[width=0.5\textwidth]{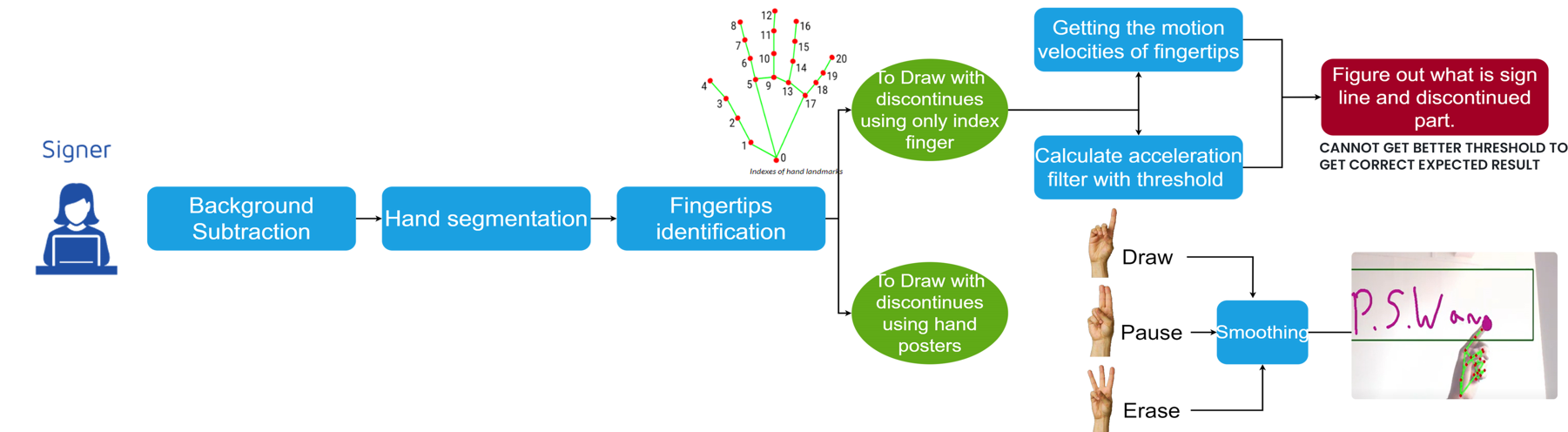}
    \caption{Flow-Chart of Signing System}
    \label{fig:flowchart}
\end{figure}

\begin{figure}[ht]
    \centering
    \includegraphics[width=0.3\textwidth]{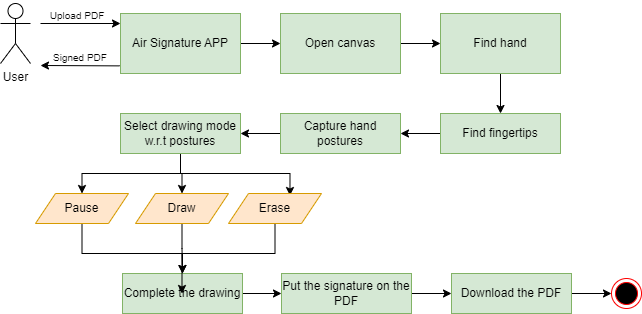}
    \caption{Air signing Block Diagram}
    \label{fig:blockdiagram}
\end{figure}

\subsection{Signature Verification}
\subsubsection{Data Collection}
To evaluate our signature verification algorithm, we have considered two widely used benchmark databases: (1) CEDAR Synthetic Signature Database, and (2) IN AIR SIGNATURE databases.

\paragraph{CEDAR}
There are 55 signers in the CEDAR signature database, all from different racial, ethnic, and professional backgrounds. These signers each put their real signature on 24 documents at 20-minute intervals. To create 24 fabricated signatures for each of the real signers, the forgers each attempted to imitate the signatures of three people, eight times each. Hence the dataset comprises 55 × 24 = 1,320 genuine signatures as well as 1,320 forged signatures. This offers grayscale versions of the signature images.

\begin{figure}[ht]
    \centering
    \includegraphics[width=0.3\textwidth]{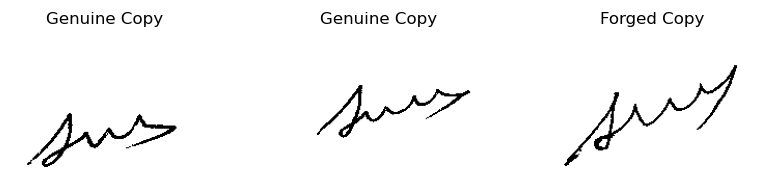}
    \caption{An example of CEDAR Data set}
    \label{fig:cedardataset}
\end{figure}

\begin{figure}[ht]
    \centering
    \includegraphics[width=0.3\textwidth]{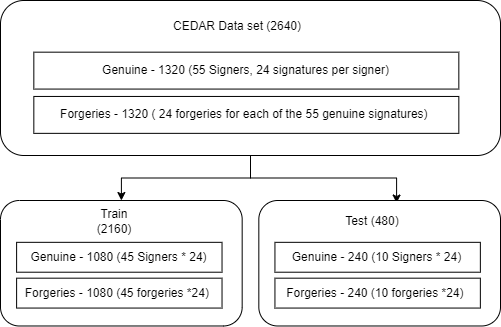}
    \caption{Train/test split based on authors, irrespective of the loss function and training data generation strategy}
    \label{fig:cedarsplit}
\end{figure}

\paragraph{IN AIR SIGNATURE DATABASES}
There are now two in-air signature databases. The creation of each of the two databases involved forty (40) volunteers. Each participant imitates five other participants' signatures while simultaneously signing five times in the air. Two methods are included in the suggested process for data collection for each of the two in-air signature databases. The participant signs in the air directly in front of the camera in the first database. Using a transparent glass plate placed between them and the camera, participants in the second database sign their names in the air. Hence the dataset comprises 40 × 10 = 400 genuine signatures as well as 400 forged signatures.

\begin{figure}[ht]
    \centering
    \includegraphics[width=0.4\textwidth]{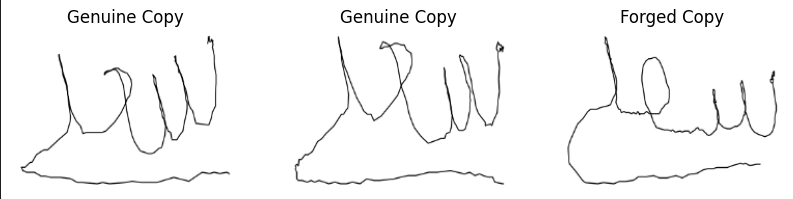}
    \caption{Forged and Genuine Signatures of IEEE In Air Signatures}
    \label{fig:ieeeinair}
\end{figure}

\subsubsection{Preprocessing}
Since batch training in a neural network normally requires photos of the same size, the size range of the signature images we are considering is from 153 × 258 to 819 × 1137. Bi-linear interpolation is used to resize each image to a set size of 155 × 220. Afterward, we invert the images so that the background pixels have 0 values. Furthermore, we divide the pixel values in each image by 255 to normalize it, hence all pixel values become between 0 to 1.

\subsubsection{Siamese Network and Architecture}
A family of network designs known as Siamese Neural Networks (SNN) often have two identical sub-networks. The identical configuration of the twin CNN includes the same parameters and shared weights. Both of the sub-networks mirror the parameter updating. This method has been effectively utilized for face recognition \cite{b18}.

\begin{figure}[ht]
    \centering
    \includegraphics[width=0.7\linewidth]{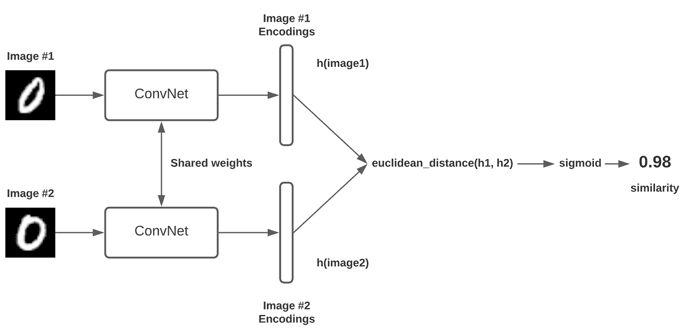}
    \caption{Siamese Basic Network Architecture}
    \label{fig:siamesearchitecture}
\end{figure}

These sub-networks are joined by a loss function at the top, which computes a similarity metric involving the Euclidean distance between the feature representation on each side of the Siamese network. One such loss function that is mostly used in Siamese networks is the contrastive loss defined as follows:

\begin{figure}[ht]
    \centering
    \includegraphics[width=0.4\linewidth]{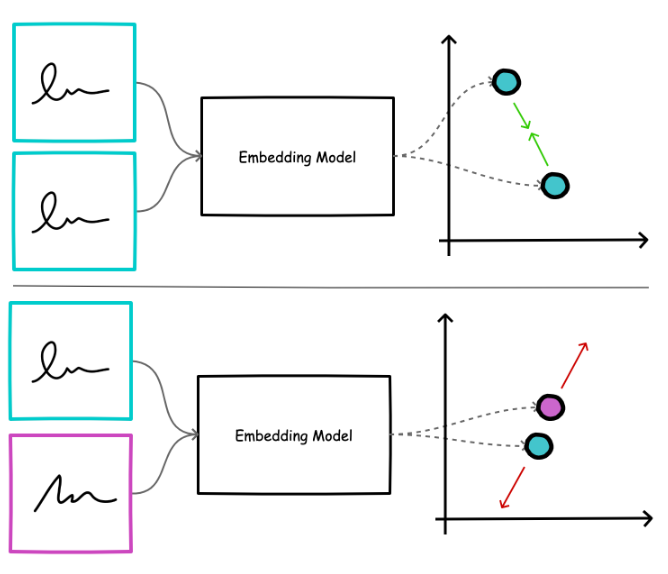}
    \caption{SNN contrastive loss}
    \label{fig:lossfunction}
\end{figure}

\begin{center}
$ \text{Loss} = 
\begin{cases} 
D & \text{if pair is positive} \\ 
\max(0, m-D) & \text{if pair is negative} 
\end{cases} $
\end{center}

where \textit{D} is the calculated distance between embeddings for each data point in the pair, and \textit{m} is a constant value of margin. The contrastive loss function is set up such that we minimize the distance between embeddings for positive pairs, and maximize the distance between embeddings for negative pairs.

The Siamese network's branches can each be thought of as functions that embed the input image in a space. This space will have the property that pictures of the same class (a genuine signature for a particular writer) will be closer to each other than images of various classes (forgeries or signatures from various writers) due to the loss function chosen. A layer that calculates the Euclidean distance between the two points in the embedded space connects the two branches. The next step is to establish a threshold value for the distance between two photos to assess whether they belong to the same class (genuine, genuine) or a distinct class (genuine, forged).

The initial convolutional layers apply 96, 11 x 11 kernels with a 1-pixel stride to the 155 x 220 input signature image to filter it. The output of the first convolutional layer, which has been response-normalized and pooled, serves as the input for the second convolutional layer, which applies 256 kernels of size 55 to filter it. Without any pooling or layer normalization interventions, the third and fourth convolutional layers are connected. The third layer has 384 kernels of size 3 × 3 connected to the (normalized, pooled, and dropout) output of the second convolutional layer. The fourth convolutional layer contains 256, 3 × 3 kernels. As a result, the neural network learns more higher-level or more abstract characteristics while learning fewer lower-level features for larger receptive fields. 1024 neurons make up the first fully linked layer, whereas 128 neurons make up the second fully connected layer. The highest learned feature vector from either side of the network has a dimension of 128 according to this.

\begin{figure}[ht]
    \centering
    \includegraphics[width=0.8\linewidth]{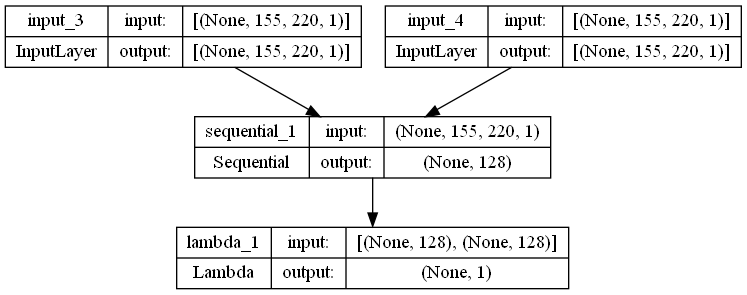}
    \caption{Model Architecture}
    \label{fig:modelarchitecture}
\end{figure}

We trained the model using RMSprop for 100 epochs, using a momentum rate equal to 0.9, and mini-batch size equal to 128. We started with an initial learning rate (LR) equal to 1e-4 with hyperparameters $\rho$ = 0.9 and $\epsilon$ = 1e-8. With TensorFlow serving as the backend, the Keras library is used across our entire framework. According to several databases, the training took nearly 10 hours to complete using a GeForce GTX 1070.

\begin{figure}[ht]
    \centering
    \includegraphics[width=0.7\linewidth]{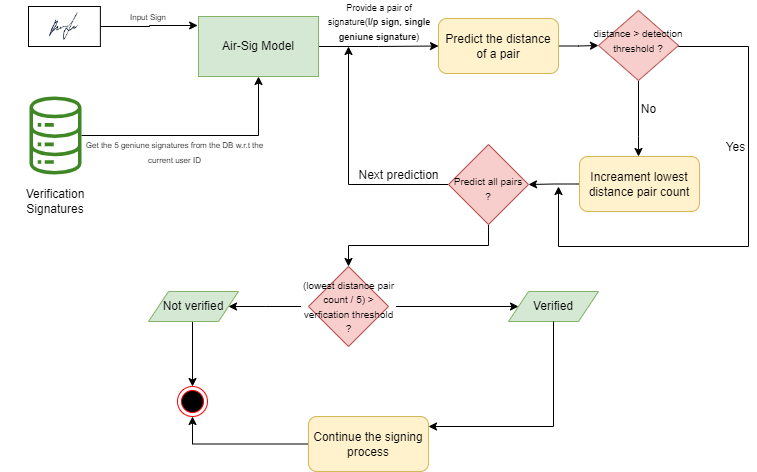}
    \caption{Verification Model Architecture}
    \label{fig:verificationarchitecture}
\end{figure}

\section{Results and Discussions}
Figure \ref{fig:flowchart} shows our Air signing process. In this process, we first detect the human hands then through our method we detect the fingertip of the hand. The user can draw the signature inside that box in the video-capturing canvas. After the user finished drawing the signature press the \textit{Q} or \textit{ENTER} to exit and save that image in the application. We test with different lighting and some complex background but it is sometimes difficult to detect the hands in those situations. Finally, we tried different types of postures to check whether our application correctly works with our defined postures to control drawing while signing.

\begin{figure}[ht]
    \centering
    \includegraphics[width=0.2\textwidth]{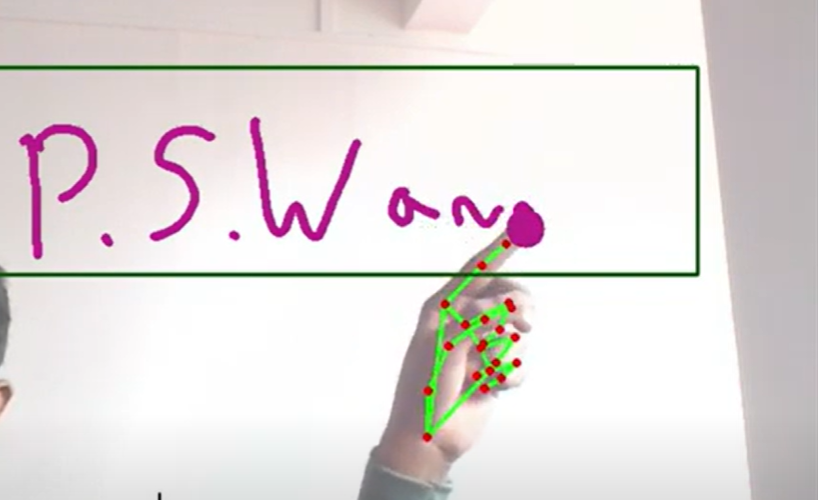}
    \caption{Air Signing output}
    \label{fig:airsigningoutput}
\end{figure}

For Signature Verification we train and test our model with selected two data sets. Those final test accuracies are shown in Table \ref{tab:comparison}. As shown in the table the model trained on the \textit{CEDAR} data set gives good accuracy, so we selected that model to verify our user's signatures with at least one positive signature image of that user.

\begin{table}[htbp]
\caption{Comparison of the proposed method on two signature databases}
\begin{center}
\begin{tabular}{|c|c|c|}
\hline
\textbf{}&\multicolumn{2}{|c|}{\textbf{Datasets}} \\
\cline{2-3} 
\textbf{} & \textbf{\textit{CEDAR}}& \textbf{\textit{In Air IEEE}}\\
\hline
\textbf{Accuracy} & \textbf{\textit{0.871}}& \textbf{\textit{0.528}}\\
\hline
\textbf{FAR} & \textbf{\textit{5.39\%}}& \textbf{\textit{2.5\%}}\\
\hline
\textbf{FRR} & \textbf{\textit{7.48\%}}& \textbf{\textit{15\%}}\\
\hline
\end{tabular}
\label{tab:comparison}
\end{center}
\end{table}

\begin{figure}[ht]
    \centering
    \includegraphics[width=0.4\textwidth]{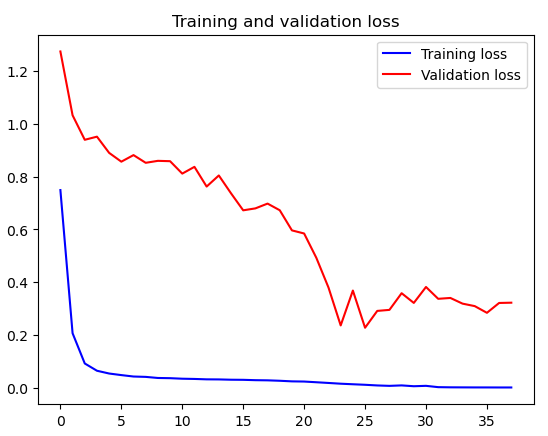}
    \caption{Training and Validation Loss vs Epoch}
    \label{fig:lossgraph}
\end{figure}

As shown in Fig. \ref{fig:lossgraph}, both training and validation losses continuously fall over every epoch. But at some epochs, it finished early stopping because, over some continuous epochs, there is no fall in validation loss. So we loaded the weights from the epoch which gave the best validation accuracy during the training period.

\begin{figure}[ht]
    \centering
    \includegraphics[width=0.5\textwidth]{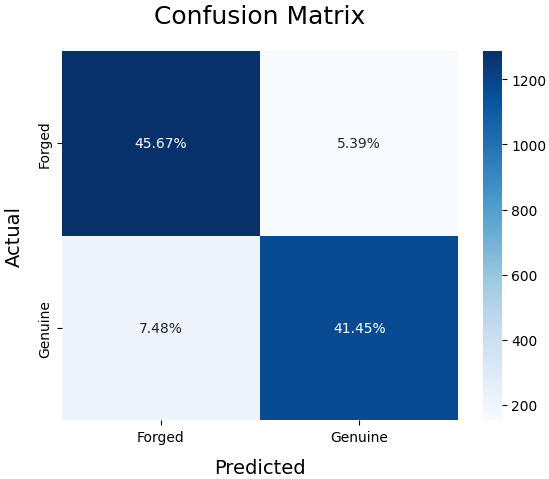}
    \caption{Confusion Matrix for \textit{CEDAR} Test Set}
    \label{fig:confusionmatrix}
\end{figure}

As shown in Fig. \ref{fig:confusionmatrix}, the Confusion Matrix False Acceptance Rate (FAR) is relatively low (5.39\%), but for this kind of security-based application, it needs to be ideally zero. The False Rejection Rate (FRR) is also relatively low (7.48\%) compared to the accuracy, but it is not the case in this application because it will predict genuine as forged so it is secure in this application.

\begin{figure}[ht]
    \centering
    \includegraphics[width=0.4\textwidth]{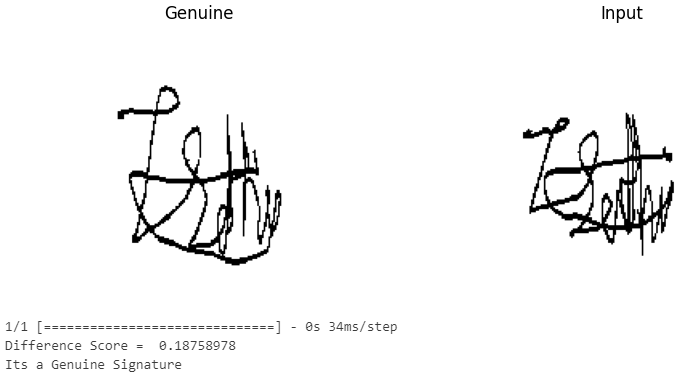}
    \caption{Prediction on pair of Positive and Input}
    \label{fig:prediction}
\end{figure}

\section{Conclusion}
The importance of digital signing on soft copies is a fundamental aspect of the contemporary online environment \cite{b1}. To enhance the effectiveness and user-friendliness of device interactions, there is a need to explore non-contact video-based interaction methods. In this research, we propose a robust technique for fingertip detection, leveraging MediaPipe for accurate fingertip tracking even in complex environments. Subsequently, the user can perform a signature gesture in front of the camera using their fingertip, with the resulting signature securely stored. To protect documents from unauthorized forgery attempts, we employ Siamese Neural Networks (SNN) \cite{b18} as part of the neural cores. The SNN utilizes a machine learning algorithm to match air signatures with the individual's stored signatures before digitally signing the PDF, ensuring document integrity \cite{b1}. This approach offers a secure and efficient method of verification. In addition, our plans involve the development of an intelligent end-to-end solution for natural-style signing, enabling users to authorize themselves using their signatures across various online transactions and mobile login processes.

\balance
\end{document}